\documentclass[journal]{IEEEtran}
\usepackage{mathrsfs}
\usepackage{amsfonts,amssymb}
\usepackage{amsmath}
\usepackage{CJK}
\usepackage{cases}
\usepackage{bm}
\usepackage{float}
\usepackage{multicol}
\usepackage{graphicx}
\usepackage{subfigure}
\usepackage{hyperref}

\usepackage{array}
\usepackage{algorithm}
\usepackage{algpseudocode}
\usepackage[mathscr]{euscript}
\usepackage{dblfloatfix}
\usepackage{etoolbox}

\ifCLASSINFOpdf
\else
\fi
\makeatletter
\def\endthebibliography{%
  \def\@noitemerr{\@latex@warning{Empty `thebibliography' environment}}%
  \endlist
}
\makeatother
\begin{document}
%
\title{Qualitative Measurements of Policy Discrepancy for Return-Based Deep Q-Network}
%
%

\author{Wenjia~Meng,~Qian~Zheng,~Long~Yang,~Pengfei~Li, and Gang~Pan 
\thanks{\textsuperscript{\textcopyright} 20xx IEEE. Personal use of this material is permitted. Permission from IEEE must be obtained for all other uses, in any current or future media, including
reprinting/republishing this material for advertising or promotional purposes, creating new collective works, for resale or redistribution to servers or lists, or reuse of any copyrighted component of this work in other works. 
The final version can be available online at http://ieeexplore.ieee.org and its Digital Object Identifier is  10.1109/TNNLS.2019.2948892.}
}

%



\maketitle

\begin{abstract}

The deep Q-network (DQN) and return-based reinforcement learning are two promising algorithms proposed in recent years. DQN brings advances to complex sequential decision problems, while return-based algorithms have advantages in making use of sample trajectories. 
In this paper, we propose a general framework to combine DQN and most of the return-based reinforcement learning algorithms, named R-DQN.
We show the performance of traditional DQN can be significantly improved by introducing return-based algorithms. 
In order to further improve the R-DQN, we design a strategy with two measurements to qualitatively measure the policy discrepancy. 
We conduct experiments on several representative tasks from the OpenAI Gym and Atari games. 
The state-of-the-art performance achieved by our method with this proposed strategy validates its effectiveness.  
\end{abstract}


\IEEEpeerreviewmaketitle

\section{Introduction}
Reinforcement learning has achieved impressive performance on sequential decision problems \cite{sutton1998reinforcement}, \cite{SuttonMSM99}, \cite{MnihKSRVBGRFOPB15}. The most recent successful reinforcement learning method is deep Q-network (DQN) \cite{MnihKSRVBGRFOPB15}, which combines Q-learning with a deep neural network. 
It kick-starts recent advances in complex sequential decision-making problems. 
The success of DQN largely benefits from experience replay \cite{Lin92}, which enables it to perform the update from samples \cite{MnihKSRVBGRFOPB15}. 
However, the traditional DQN is a bootstrap method which only makes use of one-step samples. 
Such traditional DQN is not stable enough with function approximation \cite{MunosSHB16}. 
A promising approach to address such unstability is to combine DQN with return-based algorithms that learn from sampled multi-step returns \cite{MunosSHB16}, \cite{WangBHMMKF16}, \cite{gruslys2017reactor}.

These works \cite{MunosSHB16}, \cite{WangBHMMKF16}, \cite{gruslys2017reactor} improve the training model by merging the DQN with a new formulation of returns. 
However, the formulations of returns in these works are specialized, resulting in less feasibility to make full use of different return-based algorithms (\emph{e.g.}, \cite{Watkins1992}, \cite{PengW96}, \cite{vanHasselt11}). 
Moreover, the qualitative policy discrepancy is not explicitly considered in these works, which is beneficial to utilizing off-policy returns \cite{HarutyunyanBSM16}. 
In order to address these two issues, we propose a unified formulation that can combine the DQN and most return-based algorithms and study to qualitatively measure policy discrepancy in this unified framework. 
Specifically, in the proposed unified formulation, we adopt off-policy corrections to correct returns since original return-based methods are on-policy and do not work well with experience replay \cite{MunosSHB16}, \cite{HarutyunyanBSM16}. 
Such off-policy corrections are based on trace coefficients which are used to calculate the utilization level of returns \cite{MunosSHB16}. 
These coefficients can correct policy discrepancy \cite{MunosSHB16} which represents the discrepancy between target policy $\pi$ and behavior policy $\mu$. 
It is noticeable that $\pi$ and $\mu$ represent the policy being learned about and the policy used to generate behavior respectively \cite{sutton1998reinforcement}. 
In the proposed unified framework, we further study qualitative policy discrepancy, qualitative classification about whether target and behavior policies are similar or not, to enable the trace coefficient to automatically achieve a reasonable value in \emph{near on-policy case} and \emph{near off-policy case} (defined in \ref{subsec:3.2}) \cite{HarutyunyanBSM16}. 
Our contributions can be summarized as follows: 
\begin{itemize}
    \item We propose a general framework for R-DQN. With such a framework, most of the return-based reinforcement learning algorithms can be combined with DQN.
    \item We present a strategy with two measurements to qualitatively measure the policy discrepancy under our R-DQN framework. 
    \item We show by experiments that the performance of the existing DQN method can be significantly improved with the proposed R-DQN framework and the state-of-the-art performance can be achieved with the proposed strategy. 
\end{itemize}

\begin{table*}
\vspace{-25pt}
\centering
\caption{Summary of return-based algorithms}\label{tab:table1}
\vspace{-5pt}
\begin{tabular}{ l | c c c } 
\hline
Algorithm         & $Z(x')$ & $\delta_t$ & $C_s$  \\
\hline
\hline
Watkins's Q($\lambda$)  & $\mathop {\max }\limits_a Q({x'},a)$  & ${r_{t}} + \gamma \mathop {\max }\limits_a (Q({x_{t + 1}},a)) - Q({x_t},{a_t})$  & $\lambda$ \\
\hline
P $\&$ W's Q($\lambda$)    & $\mathop {\max }\limits_a Q({x'},a)$  & ${r_{t}} + \gamma \mathop {\max }\limits_a (Q({x_{t + 1}},a)) - \mathop {\max }\limits_a (Q({x_{t}},a))$ & $\lambda$ \\
\hline
General Q($\lambda$)           & $\mathbb{E_{\pi}} Q({x'},\cdot)$   & ${r_{t}} + \gamma \mathbb{E_{\pi}}Q({x_{t + 1}},\cdot) - \mathbb{E_{\pi}}Q({x_{t}},\cdot)$  & $\lambda$ \\
\hline
IS         & $\mathbb{E_{\pi}}Q({x'},.)$  & ${r_{t}} + \gamma \mathbb{E_{\pi}}Q({x_{t + 1}},\cdot) - Q({x_{t}},a_t)$ & $\frac{{\pi ({a_s}|{x_s})}}{{\mu ({a_s}|{x_s})}}$  \\
\hline
TB($\lambda$)     & $\mathbb{E_{\pi}}Q({x'},\cdot)$  & ${r_{t}} + \gamma \mathbb{E_{\pi}}Q({x_{t + 1}},\cdot) - Q({x_{t}},a_t)$ & $\lambda\pi ({a_s}|{x_s})$  \\
\hline
${Q^\pi }(\lambda )$     & $\mathbb{E_{\pi}}Q({x'},\cdot)$  & ${r_{t}} + \gamma \mathbb{E_{\pi}}Q({x_{t + 1}},\cdot) - Q({x_{t}},a_t)$ & $\lambda$  \\
\hline
Retrace($\lambda$)     & $\mathbb{E_{\pi}}Q({x'},\cdot)$  & ${r_{t}} + \gamma \mathbb{E_{\pi}}Q({x_{t + 1}},\cdot) - Q({x_{t}},a_t)$ & $\lambda \min (1,\frac{{\pi ({a_s}|{x_s})}}{{\mu ({a_s}|{x_s})}})$ \\
\hline
Our QM($\lambda$)     & -  & - & 
$
\begin{cases}
\lambda \min (1,\frac{{\pi ({a_s}|{x_s})}}{{\mu ({a_s}|{x_s})}}),       & \text{near on-policy} \\
\lambda \pi ({a_s}|{x_s}),   & \text{near off-policy}
\end{cases}
$\\
\hline
\end{tabular}
\vspace{-15pt}
\end{table*}

\vspace{-10pt}
\section{Related Work}\label{sec:2}

As we focus on combining return-based algorithms and deep Q-network in this paper, we briefly review these two components. 
\vspace{-8pt}
\subsection{Return-based Algorithms}\label{section II-A}
Return-based algorithms are effective on estimating value function in reinforcement learning \cite{sutton1998reinforcement}, \cite{BartoD93}. 
Original return-based methods which are on-policy are always considered as less effective on tasks with experience replay \cite{MnihKSRVBGRFOPB15}. 
Off-policy corrections are widely used to address this problem \cite{MunosSHB16}, \cite{HarutyunyanBSM16}. 
Such off-policy corrections can be regarded as an approach to combine on-policy and off-policy methods \cite{GuLGTL16}, \cite{GuLTGSL17}.  
The off-policy corrections in return-based algorithms depend on policy discrepancy. 
Most existing return-based algorithms can be divided into four categories according to the degree of dependency on policy discrepancy. 
In the following, we briefly introduce these return-based algorithms by category. 

The return-based algorithms in the first category do not take policy discrepancy into account at all, \emph{e.g.},  P$\&$W's Q($\lambda$) \cite{PengW96}, \cite{PengW93} and General Q($\lambda$) \cite{vanHasselt11}. 
The return-based algorithms in the second category implicitly consider policy discrepancy by correcting rewards, \emph{e.g.}, ${Q^\pi (\lambda )}$ and ${Q^* }(\lambda )$ \cite{HarutyunyanBSM16}. 
The return-based algorithms in the third category explicitly consider policy discrepancy in terms of target policy, \emph{e.g.}, Watkins's Q($\lambda$) \cite{Watkins1992} and TB($\lambda$) (Tree-backup) \cite{PrecupSS00}. 
The return-based algorithms in the fourth category explicitly take policy discrepancy into account in terms of the likelihood ratio between target policy and behavior policy, \emph{e.g.}, IS (Importance Sampling) \cite{PrecupSS00} and Retrace($\lambda$) \cite{MunosSHB16}. 

Even though these return-based algorithms consider policy discrepancy in varying degrees, they do not address qualitative policy discrepancy which can enable algorithms to efficiently benefit from returns by distinguishing near on-policy case from near off-policy case \cite{HarutyunyanBSM16}. 
\vspace{-8pt}
\subsection{Deep Q-network}
The deep Q-network can provide rich representations of the environment to perform well \cite{MnihKSRVBGRFOPB15}, \cite{gruslys2017reactor}, \cite{SilverHMGSDSAPL16}, \cite{LillicrapHPHETS15}, \cite{SchulmanLAJM15}. 
Many extensions have been proposed to enhance its speed or stability. 
Here, we introduce several representative works among these extensions. 

Double DQN \cite{HasseltGS16} addresses the overestimation issue of DQN by decomposing action selection and action evaluation. 
The dueling network architecture \cite{WangSHHLF16} can generalize learning across actions to achieve better policy evaluation by utilizing the dueling network to separately represent state value function and advantage function. 
Prioritized experience replay \cite{SchaulQAS15} can replay important transitions more frequently to enable DQN to learn more efficiently. 
Rainbow \cite{HesselMHSODHPAS18} achieves the state-of-the-art results by integrating the ideas of different DQN algorithms. 
Safe and efficient off-policy reinforcement learning \cite{MunosSHB16} can enable deep Q-network to benefit from specific return by combining deep Q-network with Retrace($\lambda$). 
However, these works do not focus on the general return and the sample trajectory is not efficiently well used.
\vspace{-8pt}
\section{A Unified Return-based Update Target}\label{sec:3}

In order to combine the general return-based algorithm with deep Q-network, we propose a unified return-based update target in this section. Based on this unified update target, we give two definitions of near on- and off-policy cases and propose a new strategy, called QM($\lambda$), to qualitatively measure policy discrepancy.

The proposed unified return-based update target $Y(x, a)$ for action value $Q(x, a)$ is formulated as:
\vspace{-5pt}
\begin{equation}\label{equation 1}
        Y(x,a) = {r_0} + \gamma Z(x') +\mathbb{E_{\mu}}\left[ {\sum\limits_{t \ge 1}^{} {{\gamma ^t}(\prod\limits_{s = 1}^t {{C_s}} ){\delta _t}} } \right]
\end{equation}
where $r_0$ represents the immediate reward, $\gamma$ is discount rate, $Z(x')$ estimates the expectation of state value for next state $x'$, $C_s$ represents non-negative trace coefficient, ${\delta _t}$ represents the temporal difference error at step $t$.

In the formulation of the unified return-based update target, the trace coefficient $C_s$ is critical as it is related to the decay of trace. 
For near on-policy case, the trace coefficient is expected to make full use of return. 
For near off-policy case, the trace coefficient is expected to efficiently cut the trace. 
In the following, we show several typical return-based algorithms corresponding to the works introduced in Section \ref{section II-A} under the proposed unified formulation (\ref{equation 1}). 
A brief summation of these algorithms is shown in Table \ref{tab:table1}. 
It is noticeable that the definitions of $Z(x')$, ${\delta _t}$ and $C_s$ in Table \ref{tab:table1} can be found in \cite{MunosSHB16}, \cite{HarutyunyanBSM16}. 

\vspace{-10pt}
\subsection{Algorithm Analysis under the Unified Formulation} \label{subsec:3.1}

\textbf{No Policy Discrepancy.}
P$\&$W's Q($\lambda$) \cite{PengW96} and general Q($\lambda$) \cite{vanHasselt11} do not consider policy discrepancy. The computation of their trace coefficient $C_s$ is not related to policy discrepancy.

\textbf{Implicit Policy Discrepancy.}
Although ${Q^\pi }(\lambda )$ and ${Q^* }(\lambda )$ \cite{HarutyunyanBSM16} does not consider policy discrepancy in terms of trace coefficient $C_s$, it implicitly takes account of policy discrepancy in terms of correcting rewards with an off-policy correction $\gamma \mathbb{E_{\pi}}Q({x_{t + 1}},\cdot) - Q({x_{t}},a_t)$.

\textbf{Policy Discrepancy in Terms of Target Policy.}
Watkins's Q($\lambda$) \cite{Watkins1992} and TB($\lambda$) \cite{PrecupSS00} explicitly consider policy discrepancy in terms of the
target policy rather than correcting rewards.
\begin{itemize}
    \item Watkins's Q($\lambda$) directly cuts the trace by setting trace coefficient $C_s$ to zero when the sampled action is not the greedy one under target policy.
    \item TB($\lambda$) adopts trace coefficient $C_s$ which is proportional to target policy probability $\pi ({a_s}|{x_s})$ to discount the trace rather than cutting it. 
\end{itemize}

\textbf{Policy Discrepancy in Terms of Importance Sampling Ratio.}
IS \cite{PrecupSS00} and Retrace($\lambda$) \cite{MunosSHB16} consider policy discrepancy by taking both target and behavior policies into account.
\begin{itemize}
    \item IS corrects policy discrepancy by setting the trace coefficient $C_s$ to be proportional to the ratio $\frac{{\pi ({a_s}|{x_s})}}{{\mu ({a_s}|{x_s})}}$ where ${\pi ({a_s}|{x_s})}$ and ${\mu({a_s}|{x_s})}$ represent target and behavior policies respectively (($x_s, a_s$) stands for a given state action pair).
    \item Retrace($\lambda$) allows consideration of policy discrepancy by utilizing an importance sampling ratio truncated at 1. 
\end{itemize}

\vspace{-15pt}

\subsection{Qualitative Policy Discrepancy: Near On- and Off-Policy Cases}\label{subsec:3.2}

Even though algorithms above consider policy discrepancy in varying degrees, they do not address qualitative policy discrepancy which can enable the trace coefficient to achieve a reasonable value on near on- and off-policy cases \cite{HarutyunyanBSM16}. Therefore, we propose a strategy that can qualitatively measure policy discrepancy, which is inspired by the formulations of the trace coefficient in the algorithms above.
Algorithms with our strategy are expected to differentiate near on-policy case and near off-policy case.

We propose two definitions for near on- and off-policy cases.
These two definitions can classify whether a transition ($x, a, r, x'$) is near on-policy case or not. 
Such definitions are according to two different views from the idea of the dueling architecture \cite{WangSHHLF16}. 
That is, focusing on state is more desirable when actions do not affect the environment, (\emph{i.e.}, when actions are irrelevant to the environment change), 
while focusing on action is more desirable when actions can affect the environment, (\emph{i.e.}, when actions are relevant to the environment change). 
This indicates that the effectiveness of emphasizing states depends on whether actions affect the environment or not.
Therefore, one of our definitions is proposed to emphasize sampled states when actions do not affect the environment, 
while the other is proposed to emphasize the sampled action when actions affect the environment. These two definitions are detailed as follows: 

\textbf{Definition 1} (\emph{near on- and off-policy cases \footnote{Researchers call it on-policy learning when target policy is the same as behavior policy, otherwise, off-policy
learning \cite{sutton1998reinforcement}.}}) Given one transition, for the sampled state, when its greedy actions under target policy and behavior policy are the same, this transition is near on-policy case. Otherwise, it is near off-policy case.

\textbf{Definition 2} (\emph{near on- and off-policy cases}) Given one transition, for the sampled state-action pair, when the sampled action under two policies is both greedy or non-greedy, this transition is near on-policy case. Otherwise, it is near off-policy case. 

With such definitions, near on- or off-policy case in return-based algorithms can be differentiated by a given bound. 
Once near on- or off-policy case can be differentiated, the return can be more accurately approximated by choosing a more reasonable trace coefficient. The trace coefficient $C_s$ in the proposed strategy, QM($\lambda$), can be expressed as: 
\begin{eqnarray*}
C_s =
\begin{cases}
\lambda \min (1,\frac{{\pi ({a_s}|{x_s})}}{{\mu ({a_s}|{x_s})}}),       &\text{near on-policy case} \\
\lambda \pi ({a_s}|{x_s}),   &\text{near off-policy case.}
\end{cases}
\end{eqnarray*}

Our strategy separately adopts the relatively large trace coefficient from Retrace($\lambda$) and the relatively small one from TB($\lambda$) for near on- and off-policy cases, \emph{i.e.}, $\min (1,\frac{{\pi ({a_s}|{x_s})}}{{\mu ({a_s}|{x_s})}}) \geq \pi ({a_s}|{x_s})$ \cite{MunosSHB16}. 
Such setting enables the algorithm to separately make full use of returns and efficiently cut the off-policy returns on these two cases. 

It should be kindly noted that $Z(x')$ and $\delta_t$ in the proposed QM($\lambda$) vary with original return-based algorithms. 
As shown in Table \ref{tab:table1}, these existing return-based algorithms can be classified into four categories according to the formulations of $Z(x')$ and $\delta_t$. 
Each return-based algorithm category corresponds to one specific QM($\lambda$) formulation. 
Specifically, QM($\lambda$) formulations based on Watkins's Q($\lambda$), P $\&$ W's Q($\lambda$) and General Q($\lambda$) are different from each other. 
The QM($\lambda$) formulations based on IS, TB($\lambda$), $Q^{\pi}(\lambda)$ and Retrace($\lambda$) are the same. 
For the QM($\lambda$) formulations based on IS, TB($\lambda$), $Q^{\pi}(\lambda)$ and Retrace($\lambda$), 
their trace coefficients $C_s \in \lbrack 0, \frac{{\pi ({a_s}|{x_s})}}{{\mu ({a_s}|{x_s})}}  \rbrack$  ensure their update targets $Y(x, a)$ in formulation (\ref{equation 1}) converge to the estimated values, \emph{i.e.},  the value function for a policy $\pi$ ($Q^{\pi}$) and the optimal value function ($Q^{*}$)  \cite{MunosSHB16}.

\vspace{-5pt}
\section{R-DQN Framework with Two Measurements}\label{sec:4}
 
As introduced in previous sections, existing works \cite{MunosSHB16}, \cite{WangBHMMKF16}, \cite{gruslys2017reactor} which combine DQN with return-based algorithms cannot fully benefit from general return-based algorithms (\emph{e.g.}, Watkins's Q($\lambda$), P$\&$W's Q($\lambda$) and General Q($\lambda$)). 
Therefore, in this section, we propose an R-DQN framework which can combine deep Q-network with the proposed unified return-based update. 
We first describe how to combine DQN with the general return-based algorithms. 
Under this R-DQN framework, we then propose two measurements and give their bounds to qualitatively classify near on- and off-policy cases for our QM($\lambda$). 
The whole algorithm of our R-DQN is summarized in Algorithm~\ref{alg1}.  

\vspace{-8pt}
\subsection{The Proposed R-DQN} 

\begin{figure}[htb]
\vspace{-10pt}
\centering
\caption{The framework of R-DQN. Given replay memory and deep Q-network, we can obtain the proposed unified return-based update target $Y(x_t,a_t)$. This target can be used to compute the loss of return-based deep Q-network.}
\includegraphics[width=3in]{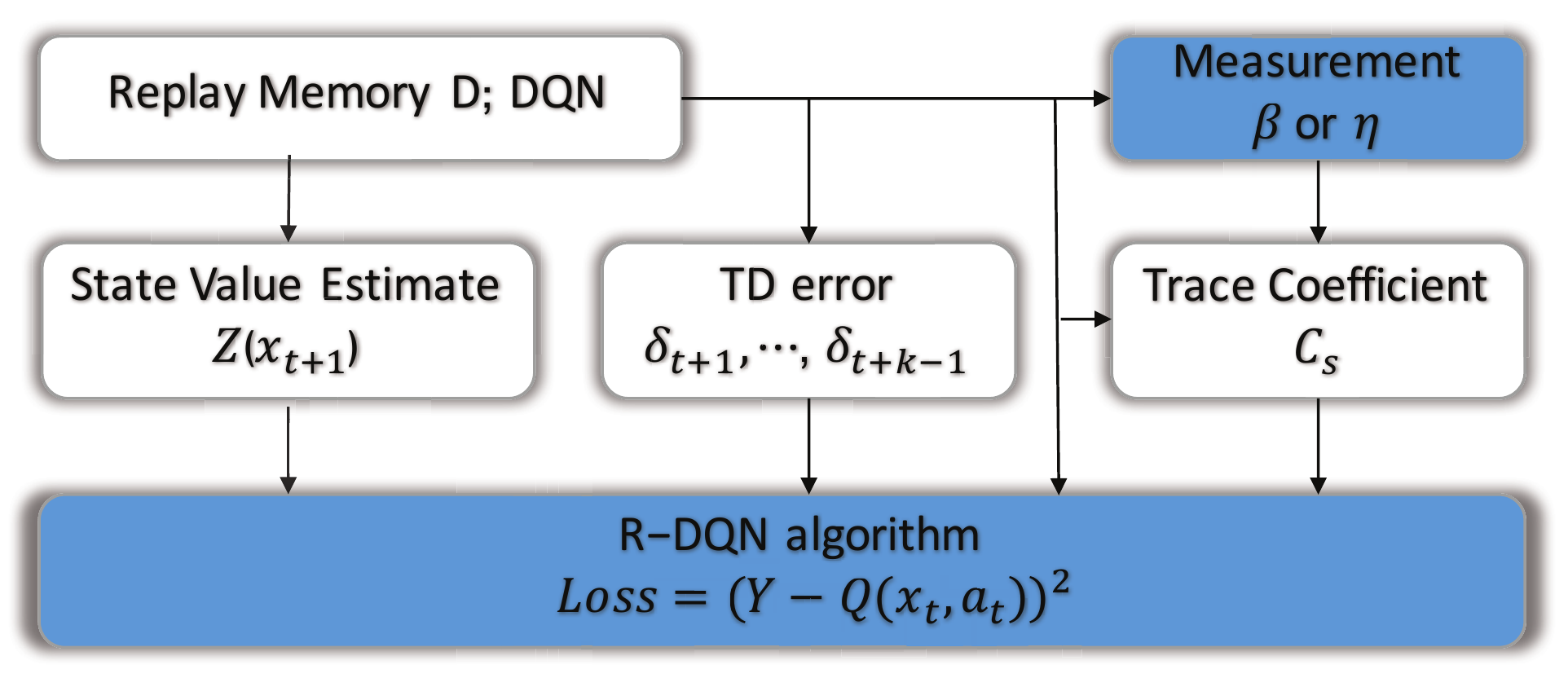}
\label{Figure 1}
\vspace{-20pt}
\end{figure}

The pipeline of R-DQN algorithms is illustrated in Figure~\ref{Figure 1}.
As shown in this figure, transitions $(x_t, a_t, r_t, x_{t+1}, \cdots, x_{t+k})$ are drawn from replay memory D. The transition sequences are utilized by R-DQN to compute state value estimate and temporal difference error. The loss can be formulated:
\vspace{-4pt}
\begin{flalign*}
    {L}({\theta_j}) ={(Y(x_t,a_t) - Q(x_t,a_t;{\theta _j}))^2}
\vspace{-10pt}
\end{flalign*}
where $\theta_j$ represents the parameters of R-DQN at step $j$. 
In the formulation, $Y(x_t,a_t)$ is represented as: 
\begin{flalign*}
Y(x_t,a_t) = r({x_t},{a_t}) + \gamma Z(x_{t+1}) + \sum\limits_{s = t + 1}^{t + k - 1} {{\gamma ^{s - t}}(\prod\limits_{i = t + 1}^s {{C_i}} ){\delta _s}} 
\end{flalign*}
\noindent where $k$ represents the number of transitions. When updating R-DQN, corresponding gradient descent is performed:
\[{\nabla _{{\theta_j}}}{L}({\theta_j}) = (Y(x_t,a_t) - Q(x_t,a_t;{\theta _j})){\nabla _{{\theta _j}}}Q(x_t,a_t;{\theta _j}).\]

\begin{algorithm}{\footnotesize}
\caption[font={tiny}]{Return-based algorithm for DQN}
\label{alg1}
\begin{algorithmic}[1]
\small
\Require Replay memory $D$ with capacity $N$; Action-value function $Q$ with random weights $\theta$
\Require Target action-value function ${\hat Q}$ with weights ${\theta ^ - } = \theta$; Replacement frequency $F$ of target network
\For{episode $= 1, M$} \label{trainQuantizationBegin}
    \State Initialize state sequence ${x_1} = \{\mathbf{s}_1\}$ 
\For{step $j = 1, T$}
    \State With probability $\epsilon$ select a random action $a_j$
    \State otherwise select action ${a_j} = \mathop {\arg \max }_{a} Q({x_j},a;\theta_j)$
    \State Execute $a_j$ in emulator, observe reward $r_j$, state ${\mathbf{s}_{j+1}}$
    \State Set state $x_{j+1} \leftarrow {\mathbf{s}_{j+1}}$ and add transition tuple $(x_j$, $a_j$, $r_j$, $x_{j+1})$ to $D$ and store behavior policy $\mu(\cdot|x_j)$ to $D$ 
    \State Randomly sample a minibatch of sequential transitions from $D$: $({x_t, a_t, r_t, ... x_{t+k}})$ and behavior policy sequences $\mu_t, \cdots, \mu_{t+k-1}$
    \State Compute $Y(x_t,a_t) = r({x_t},{a_t}) + \gamma Z({x_{t+1}}) + \sum\limits_{s = t + 1}^{t + k - 1} {{\gamma ^{s - t}}(\prod\limits_{i = t + 1}^s {{C_i}} ){\delta _s}}$ 
    \State Perform a gradient descent step on $(Y(x_t,a_t)-Q(x_t,a_t; \theta_j))^2$ with respect to the network parameters $\theta_j$

    \State Every $F$ steps reset $\hat Q = Q$
\EndFor
\EndFor \label{trainQuantizationEnd}
\end{algorithmic}
\end{algorithm}

Replay experience \cite{MnihKSRVBGRFOPB15}, \cite{Lin93} is adopted in R-DQN. There are two differences between R-DQN and DQN in experience replay: 1) 
Given state $x$, behavior policy $\mu(\cdot|x)$ is stored in R-DQN. 2) Samples drawn from replay memory in R-DQN framework are sequential.

\subsection{Qualitative Measurements and Their Corresponding Bounds in R-DQN}\label{section:iv-b}
The definition 1 and 2 in Section \ref{subsec:3.2} semantically define near on- and off-policy cases, but they cannot directly formulate these two cases. 
Some works \cite{MunosSHB16}, \cite{HarutyunyanBSM16} give more intuitive expression of these two cases. More specifically, when behavior policy is similar to target policy, the case is regarded as near on-policy; otherwise, the case is near off-policy. Therefore, we give the formulated expression of near on-policy case and near off-policy case:
\vspace{-5pt}
\begin{eqnarray*}
\text{case} =
\begin{cases}
\text{near on-policy},       & \text{measurement} < \text{bound} \\
\text{near off-policy},   & \text{measurement} >= \text{bound}
\end{cases}
\end{eqnarray*}\vspace{-10pt}

\noindent where measurement represents the dissimilarity between behavior policy and target policy, bound is used to qualitatively classify these two cases.  

In this section, inspired by the idea of \lq off-policy-ness\rq  ~\cite{HarutyunyanBSM16}, we define two measurements for policy discrepancy, namely $\beta$-based measurement and $\eta$-based measurement (as shown in Figure~\ref{Figure 1}) under our R-DQN framework. 
These two measurements are proposed according to the two definitions in Section \ref{subsec:3.2}. As these two definitions, $\beta$-based measurement and $\eta$-based measurement are proposed for emphasizing state and sampled action in their corresponding specific environments accordingly. 
Motivated by the simplicity of $L_1$ distance to measure policy discrepancy as in some of the previous works \cite{MunosSHB16}, \cite{HarutyunyanBSM16}, we adopt such metric to derive these measurements' formulations. 
Their formulations are as follows: 
\vspace{-5pt}
\begin{flalign}
    \label{eq1}
    \beta &\mathop  = \limits^{def} ||\pi(.|x_t) - \mu(.|x_t)||_1    \\
    \label{eq2}
    \eta &\mathop  = \limits^{def} |\pi(a_t|x_t) - \mu(a_t|x_t)|
\end{flalign}
\noindent when given the sampled state $x_t$ and action $a_t$.

We then derive their corresponding bounds according to these two definitions of near on- and off-policy cases. 
The derivations for the bounds are related to the experimental exploration parameter $\epsilon$ in $\epsilon$-greedy method in DQN. 
During the derivations, we assume that the final experimental exploration parameter $\epsilon$ satisfies $0 < \epsilon \leq 1/2$.
Such parameter assumption is consistent with representative works \cite{MnihKSRVBGRFOPB15}, \cite{MunosSHB16}, \cite{HasseltGS16}, \cite{WangSHHLF16}, \cite{HesselMHSODHPAS18} (\emph{e.g.}, final maximal $\epsilon$ in \cite{MunosSHB16} is equal to $1/2$, in \cite{MnihKSRVBGRFOPB15}, \cite{HasseltGS16}, \cite{WangSHHLF16}, \cite{HesselMHSODHPAS18} satisfies $0< \epsilon < 1/2$ ). 
The exploration parameters under target and behavior policies are separately represented as $\epsilon_{\pi}$ and $\epsilon_{\mu}$. 
It should be kindly noted that $|\epsilon_{\mu} - \epsilon_{\pi}| < 1/2$ can be derived according to the final exploration parameter assumption \emph{i.e.} $0 < \epsilon \leq 1/2$. 

During the derivation of measurements' bounds, the formulations of target and behavior policies are critical. 
Here, such formulations are based on $\epsilon$-greedy policy which is a common method used in DQN algorithms \cite{gruslys2017reactor}. 
In the following, we separately give these formulations for $\beta$-based and $\eta$-based measurements. 
Specifically, these formulations are proposed for DQN agents with discrete action space whose action number is $n$. 
For $\beta$-based measurement: without loss of generality, on near on-policy case, we separately formulate target and behavior policies as $\pi(\cdot|x_t)=(1-\epsilon_{\pi}+\frac{\epsilon_{\pi} }{n},\frac{\epsilon_{\pi} }{n}, \cdots,\frac{\epsilon_{\pi}}{n})$, $\mu(\cdot|x_t)=(1-\epsilon_{\mu}+\frac{\epsilon_{\mu} }{n},\frac{\epsilon_{\mu} }{n}, \cdots,\frac{\epsilon_{\mu}}{n})$ where the greedy actions under target and behavior policies are the same one; 
on near off-policy case, we separately formulate target and behavior policies as $\pi(\cdot|x_t)=(\frac{\epsilon_{\pi} }{n},1-\epsilon_{\pi}+\frac{\epsilon_{\pi} }{n}, \cdots,\frac{\epsilon_{\pi}}{n})$, $\mu(\cdot|x_t)=(1-\epsilon_{\mu}+\frac{\epsilon_{\mu} }{n},\frac{\epsilon_{\mu} }{n}, \cdots,\frac{\epsilon_{\mu}}{n})$ where the greedy actions under these two policies are not the same. 
For $\eta$-based measurement, the formulations of these two policies can be found in the following derivation process of the bound for $\eta$-based measurement.

In the following, we first derive the bound for $\beta$-based measurement by analyzing near on-policy case and near off-policy case (definition 1 in \ref{subsec:3.2}). 
We then derive the bound for $\eta$-based measurement by analyzing these two cases (definition 2 in \ref{subsec:3.2}). 

\begin{figure*}
\vspace{-20pt}
\centering
\includegraphics[width=7in]{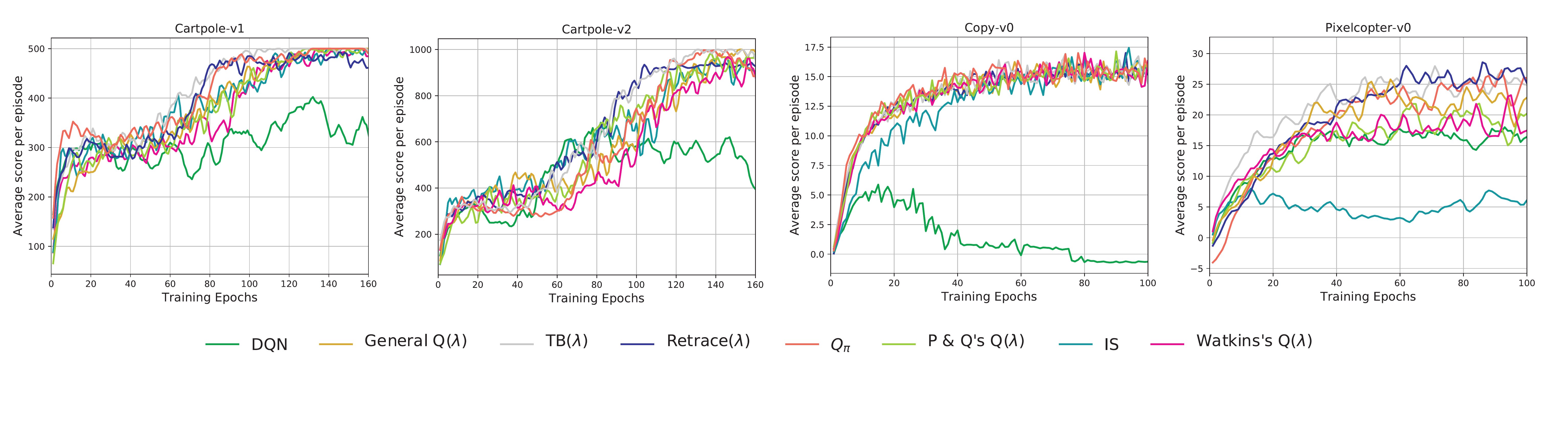}
\vspace{-8pt}
\caption{Performance comparison between R-DQN algorithms and DQN. For CartPole and Copy: the network architectures are composed of a fully connected neural network with one hidden layer of 64 neurons. For Pixelcopter: the network architecture is the same as DQN in~\cite{MnihKSRVBGRFOPB15}. 
The final exploration parameter ($\epsilon$) for R-DQN algorithm switches randomly (probability 0.3, 0.4, and 0.3 respectively) between the values (0.5, 0.1, and 0.01), which is consistent with that in \cite{MunosSHB16}. 
One epoch denotes a certain amount of environment steps. 
For CartPole-v1, CartPole-v2, Copy and Pixelcopter, one epoch respectively corresponds to 2000, 2000, 1000 and 5000 environment steps.} 

\label{Figure rdqns}
\end{figure*}

\begin{figure*}
\vspace{-10pt}
\centering
\includegraphics[width=7in]{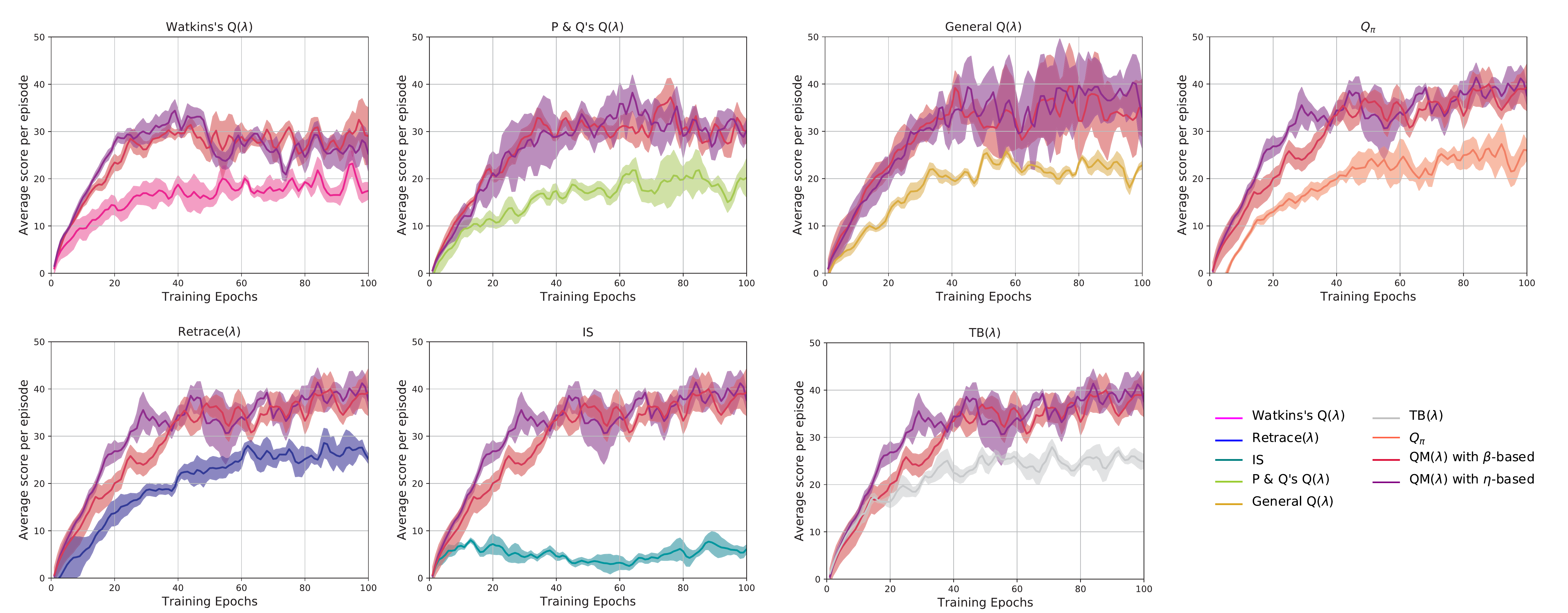}
\vspace{-10pt}
\caption{Performance comparison between R-DQN algorithms with two measurements and the ones without on Pixelcopter. The bold lines are averages over ten independent learning trials. The shaded area represents one standard deviation. `QM($\lambda$) with $\beta / \eta$-based' is the simplified representation for the method `QM($\lambda$) with $\beta/\eta$-based measurement'.}
\vspace{-20pt}
\label{Figure pixelcopter}
\end{figure*}

\begin{figure}[htb]
\centering
\includegraphics[width=3.5in]{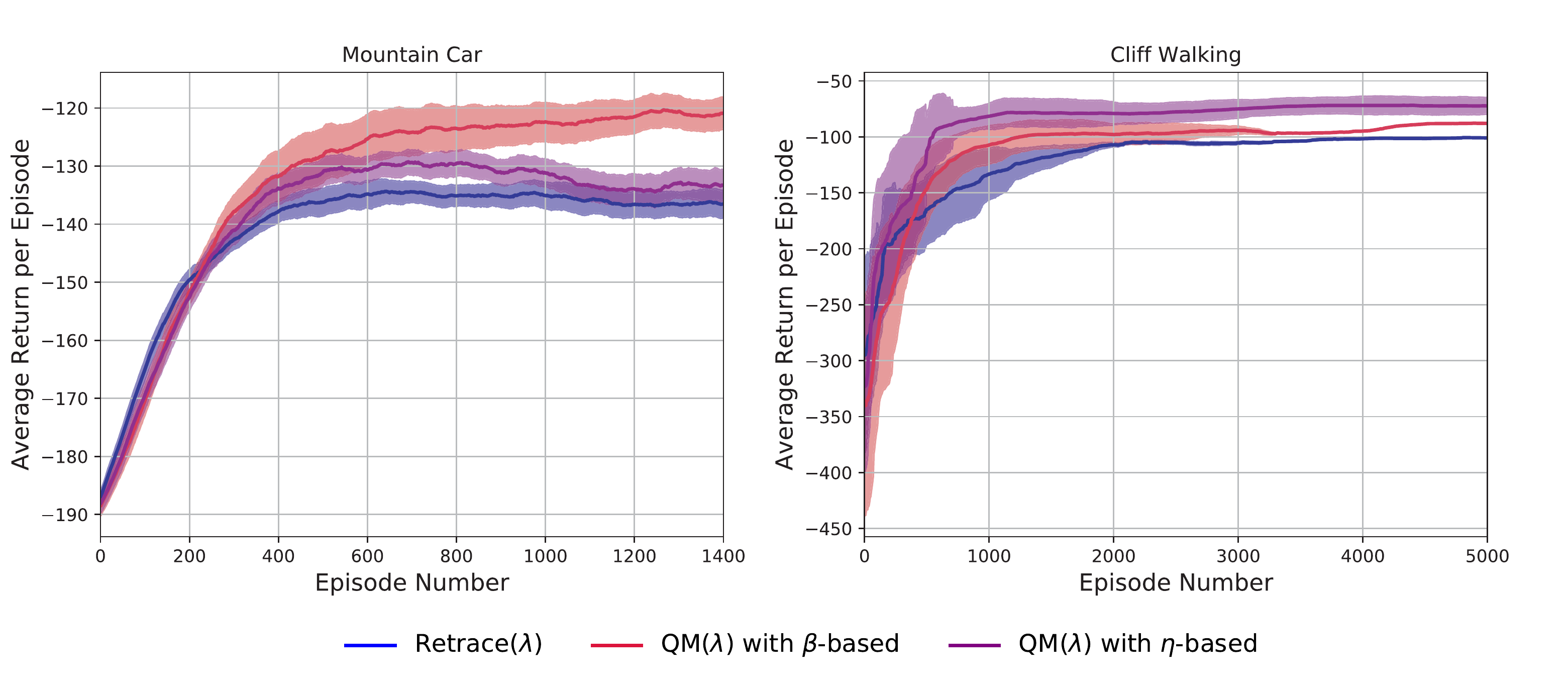}
\vspace{-20pt}
\caption{Performance comparison between QM($\lambda$) with $\beta$-based measurement and QM($\lambda$) with $\eta$-based measurement in Mountain Car and Cliff Walking. The bold lines are averages over ten independent learning trials. The shaded area represents one standard deviation.}
\vspace{-10pt}
\label{Figure mountaincar_cliff}
\end{figure}

\vspace{5pt}
\textbf{$\beta$-based measurement}
\begin{itemize}

 \item For the near on-policy case, according to the Equation (\ref{eq1}) definition for $\beta$, we can derive:
\vspace{-5pt}
  \begin{flalign*}
 \beta 
 & =||\pi(.|x_t) - \mu(.|x_t)||_1  =|\frac{\epsilon_{\mu} -\epsilon_{\pi}}{n}|(2n-2) <1. \\
 \end{flalign*}
 \vspace{-30pt}
 \item For the near off-policy case, let $\epsilon_{\pi}$ be greater than or equal to $\epsilon_{\mu}$ without loss of generality. According to the Equation (\ref{eq1}) definition for $\beta$, we can derive:
\vspace{-5pt}
\begin{flalign*}
\beta 
&=||\pi(.|x_t) - \mu(.|x_t)||_1 = 2 - {\epsilon _\pi }(\frac{2}{n}) - {\epsilon _\mu }(\frac{{2n - 2}}{n})\\
&\geq 2 - {\epsilon _\pi }(\frac{2}{n}) - {\epsilon _\pi }(\frac{{2n - 2}}{n}) = 2 - 2{\epsilon _\pi } \geq 1.\\
\end{flalign*}
\vspace{-30pt}
\item When the case is near on-policy, $\beta$ is less than $1$. Otherwise, $\beta$ is greater than or equal to $1$. We conclude that the bound of $\beta$ is $1$, under the $\epsilon$ range assumption.
\end{itemize}

\textbf{$\eta$-based measurement}

\begin{itemize}
    \item For the near on-policy case, we need to consider two situations where the sampled action under two policies is both greedy or non-greedy. In the first situation, we have
    \vspace{-15pt}
    \begin{flalign*}
        \eta &=|(1 - {\epsilon _\pi } + \frac{{{\epsilon _\pi }}}{n}) - (1 - {\epsilon _\mu } + \frac{{{\epsilon _\mu }}}{n})|\\ 
     &=|\frac{{(n - 1)({\epsilon _\mu } - {\epsilon _\pi })}}{n}| < 1/2
     \vspace{-10pt}
    \end{flalign*}\vspace{-15pt}
    
   according to the Equation (\ref{eq2}) definition for $\eta$. On the second situation, $\eta$ can be represented as $|\frac{{{\epsilon _\pi }}}{n} - \frac{{{\epsilon _\mu }}}{n}|$ which is less than $1/2$.
   
   \item For the near off-policy case, $\epsilon_\pi$ is greater than or equal to $\epsilon_\mu$ without loss of generality. We need to consider two situations where the sampled action is greedy under target policy or behavior policy. In the first situation, we can derive
  \vspace{-8pt}
  \begin{flalign*}
      |(1 - {\epsilon _\pi } + \frac{{{\epsilon _\pi }}}{n}) - \frac{{{\epsilon _\mu }}}{n}|
  &\geq |(1 - {\epsilon _\pi } + \frac{{{\epsilon _\pi }}}{n}) - \frac{{{\epsilon _\pi }}}{n}| \\
  &\geq |1 - {\epsilon _\pi }| \geq 1/2
  \end{flalign*}\vspace{-20pt}

according to the Equation (\ref{eq2}) definition for $\eta$. On the second situation, we can derive:
     \vspace{-5pt}
   \begin{flalign*}
      |(1 - {\epsilon _\mu } + \frac{{{\epsilon _\mu}}}{n}) - \frac{{{\epsilon _\pi }}}{n}|
  &\geq |(1 - {\epsilon _\pi } + \frac{{{\epsilon _\pi }}}{n}) - \frac{{{\epsilon _\pi }}}{n}| \\
  &\geq |1 - {\epsilon _\pi }| \geq 1/2.
  \end{flalign*}
 \vspace{-20pt}
 \item Based on the analysis of these two cases, we conclude that the bound of $\eta$ is $1/2$, under the $\epsilon$ range assumption. 
\end{itemize}

\begin{figure*}[htb]
\centering
    \includegraphics[width=7in]{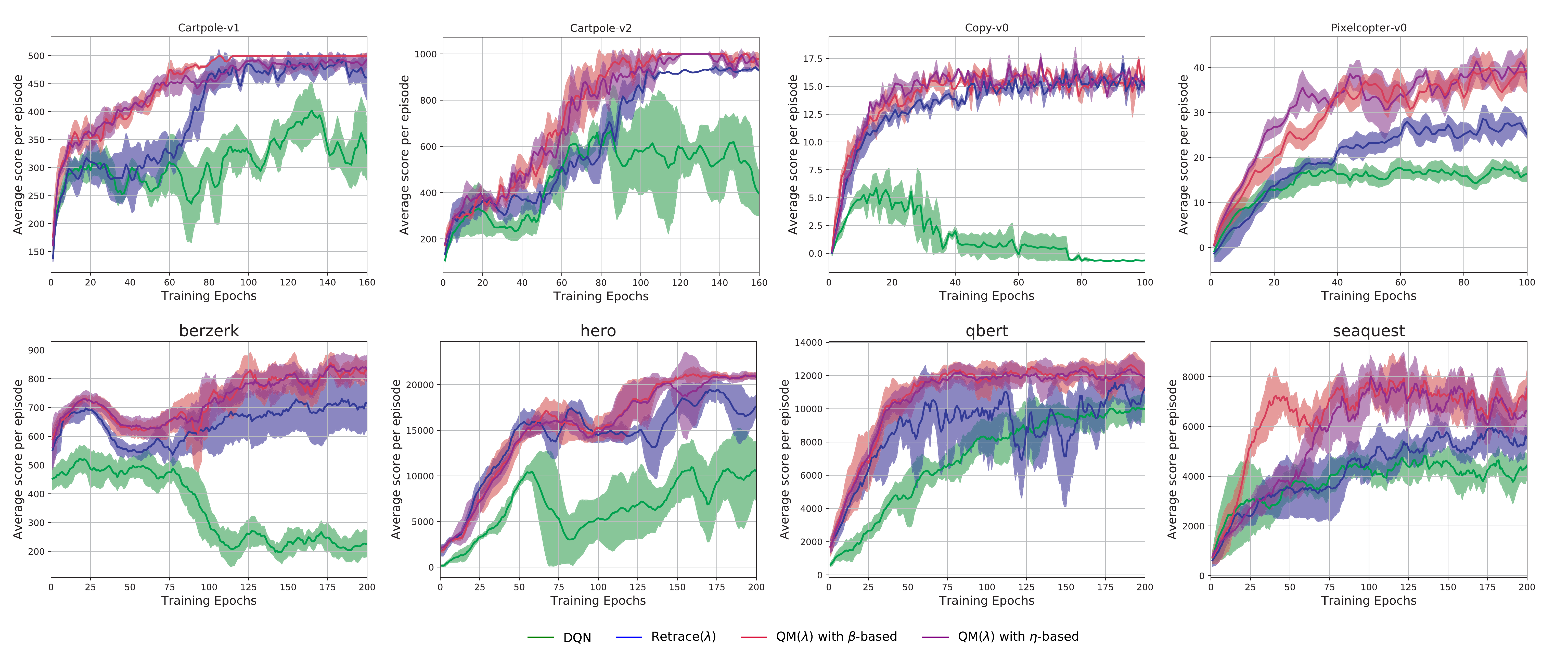}
    \vspace{-10pt}
    \caption{Performance comparison among DQN, Retrace($\lambda$), and its improved algorithm with QM($\lambda$). The performance comparison on OpenAI games and Atari games are shown on top and bottom figures respectively. The experimental configuration on Atari games is the same as that in \cite{MnihKSRVBGRFOPB15}. The bold lines are averages over ten independent learning trials. The shaded area represents one standard deviation. `QM($\lambda$) with $\beta / \eta$-based' is the simplified representation for the method `QM($\lambda$) with $\beta/\eta$-based measurement'. For these four Atari games, one epoch corresponds to 250000 environment steps.}
\vspace{-10pt}
\label{overall}
\end{figure*}

\vspace{-10pt}
\section{Experiments}\label{sec:5}

In this section, we conduct three experiments to validate the effectiveness of the proposed methods. The first experiment is to compare the R-DQN algorithms in our framework with traditional DQN. 
The second one is to validate the effectiveness of the proposed measurements. 
We compare the proposed QM($\lambda$) with the state-of-the-art method in the third experiment. We carry out experiments on three representative tasks, \emph{i.e.}, \emph{CartPole}\footnote{In this paper, we adopt two configurations (CartPole-v1, CartPole-v2) for CartPole. 
The maximum of episode steps are 500 and 1000 respectively.}, \emph{Copy}, \emph{Pixelcopter} from OpenAI Gym library~\cite{Brockman}, two classic tasks, \emph{i.e.}, \emph{Mountain Car}, \emph{Cliff Walking} in reinforcement learning and four representative Atari games, \emph{i.e.}, \emph{berzerk}, \emph{hero}, \emph{qbert}, \emph{seaquest}. 
Among these tasks, CartPole, Copy, and Pixelcopter are standard tasks of classic control, algorithmic and pygame learning environment respectively.  
All the results in Section \ref{sec:5} are averaged over ten independent learning trials with different random seeds \cite{MunosSHB16}, \cite{AnschelBS17}.
\vspace{-10pt}
\subsection{Effectiveness of R-DQN Framework}
We apply return-based algorithms under our R-DQN framework to DQN to improve its performance in this section. 
The experimental results are reported in Figure \ref{Figure rdqns}. 
As shown in Figure \ref{Figure rdqns}, we can find that all R-DQN algorithms under our R-DQN framework achieve higher scores than the traditional DQN on tasks: CartPole-v1, CartPole-v2, and Copy. 
In Figure \ref{Figure rdqns}, it can be observed that most R-DQN algorithms achieve higher or comparable scores compared to DQN in Pixelcopter. 
Specifically, in this task, IS($\lambda$)~\cite{PrecupSS00} achieves a low score as it utilizes useless trace when target policy is far away from behavior policy ($\pi(a_t|x_t) \gg \mu(a_t|x_t)$). 
The experimental results in Figure \ref{Figure rdqns} show that DQN can be improved by the proposed R-DQN framework. 
\vspace{-10pt}
\subsection{Effectiveness of the Proposed Measurements}
In order to evaluate the effectiveness of the proposed two measurements, we apply QM($\lambda$) with these two measurements to R-DQN algorithms in our framework.   
The experimental results are shown in Figure~\ref{Figure pixelcopter} and Table~\ref{tab:table2}.\footnote{It should be kindly noted that the results of QM($\lambda$) based on IS, TB($\lambda$), $Q^{\pi}(\lambda)$ and Retrace($\lambda$) are the same due to their same QM($\lambda$) formulations. 
The results of QM($\lambda$) based on Watkins's Q($\lambda$), P $\&$ W's Q($\lambda$) and General Q($\lambda$) are different due to their different QM($\lambda$) formulations.}
In the following, we separately analyze the results from Figure~\ref{Figure pixelcopter} and Table~\ref{tab:table2}. 
In Figure \ref{Figure pixelcopter}, it is noticeable that the combined R-DQN algorithms with these two measurements achieve higher scores than original R-DQN algorithms over the whole training period on Pixelcopter. 
From Figure \ref{Figure pixelcopter}, we can also observe that $\beta$-based measurement and $\eta$-based measurement can help R-DQN algorithms to achieve comparable scores. 
In Table~\ref{tab:table2}, R-DQN algorithms are classified into four categories according to the values of $\delta_t$ and $Z(x')$. 
As shown in Table \ref{tab:table2}, the improved R-DQN algorithms can achieve higher scores than original R-DQN algorithms over the four categories. 
Such results show the effectiveness of the proposed measurements. 

In order to further show the respective advantages of these two measurements, we conduct experiments on two representative samples: Mountain Car and Cliff Walking. 
The actions of states in Mountain Car hardly affect the environment, while the actions in Cliff Walking can make the agent fall into the cliff and return to the start position (more details about them can be found in \cite{sutton1998reinforcement}). The characteristics of these two tasks are well matched to $\beta$-based measurement and $\eta$-based measurement respectively.
The experimental results are shown in Figure \ref{Figure mountaincar_cliff}, QM($\lambda$) with $\beta$-based measurement achieves the highest average return in Mountain Car. In Cliff Walking, a gap of the QM($\lambda$) with $\eta$-based measurement and the other algorithms can also be observed.

\begin{table}[htb]
\centering
\vspace{-10pt}
\scriptsize
\caption{Performance comparison among all R-DQN algorithms on CartPole-v1, CartPole-v2, Copy, Pixelcopter. The reported results (averaged score $\pm$ one standard deviation) are evaluated at the end of the training epochs. `QM($\lambda$) with $\beta / \eta$' is the simplified representation for the method `QM($\lambda$) with $\beta/\eta$-based measurement'.}
\label{tab:table2}
\begin{tabular}{ l | c | c | c | c }

\hline
          Methods  & Cart1 & Cart2 & Copy & Pixel  \\
\hline\hline

DQN        &327.3 $\pm$ 50.4     &395.4 $\pm$ 93.0 & -0.6 $\pm$ 0.0   &  16.4 $\pm$ 1.7  \\
\hline
TB($\lambda$)        &494.7 $\pm$ 0.0 &958.0 $\pm$ 15.8 &14.4 $\pm$ 0.9 &24.8 $\pm$ 1.5 \\
IS       &498.7 $\pm$ 19.0      &905.2 $\pm$ 10.8 &15.9 $\pm$ 0.4 & 6.1 $\pm$ 0.9\\
$Q_{\pi}$        &489.9 $\pm$ 31.6 &885.48 $\pm$ 55.0 &15.6  $\pm$ 0.7 &26.0 $\pm$ 2.0 \\
Retrace($\lambda$)    &461.1 $\pm$ 40.7     &927.4 $\pm$ 0.0 &15.1 $\pm$ 0.5 &25.1 $\pm$ 0.9 \\

QM($\lambda$) with $\beta$ &\textbf{499.9 $\pm$ 3.2}  &\textbf{977.0 $\pm$ 19.0} & \textbf{15.9 $\pm$ 0.2}  & \textbf{39.3 $\pm$  4.9} \\
QM($\lambda$) with $\eta$  &493.2 $\pm$ 12.0  &947.5 $\pm$ 16.0 &14.9 $\pm$ 0.1    &37.5 $\pm$ 1.9\\
\hline
Watkins's Q($\lambda$)   &484.3  $\pm$  0.4   &887.0 $\pm$ 0.0  &16.3 $\pm$ 0.1  &17.4 $\pm$ 1.7 \\
QM($\lambda$) with $\beta$    &\textbf{494.9 $\pm$ 4.2}& \textbf{908.1 $\pm$ 15.1}  &16.0 $\pm$ 0.2  &\textbf{29.0 $\pm$ 5.9} \\
QM($\lambda$) with $\eta$   &493.3 $\pm$ 1.6  &904.2 $\pm$ 1.7 & \textbf{16.4 $\pm$ 0.4} &25.2 $\pm$ 3.5 \\

\hline
P $\&$ W's Q($\lambda$)      &496.7 $\pm$ 2.5      &980.4 $\pm$ 74.4 &15.8 $\pm$ 0.5 &20.2 $\pm$ 4.1 \\
QM($\lambda$) with $\beta$  & \textbf{500.0 $\pm$ 2.3} &\textbf{994.0 $\pm$ 15.1}  &15.4 $\pm$ 0.3  &29.9  $\pm$ 2.5 \\
QM($\lambda$) with $\eta$  &499.4 $\pm$ 1.2 &992.4 $\pm$ 12.1 &\textbf{16.0 $\pm$ 0.4}         & \textbf{30.2 $\pm$ 2.2} \\
\hline
General Q($\lambda$) &499.9 $\pm$ 0.0  &988.8 $\pm$ 0.0 &15.2 $\pm$ 0.3 &22.8 $\pm$ 0.8 \\

QM($\lambda$) with $\beta$ &500.0 $\pm$ 0.3 &\textbf{989.1 $\pm$ 4.0} &14.1 $\pm$ 0.2  &\textbf{34.9 $\pm$ 6.0 }  \\
QM($\lambda$) with $\eta$  &\textbf{500.0 $\pm$ 0.2}  &980.3 $\pm$ 5.2 &\textbf{15.3 $\pm$ 0.1}  &33.0 $\pm$ 6.5 \\         
\hline
\end{tabular}
\vspace{-15pt}
\end{table}

\subsection{Effectiveness of the Proposed QM($\lambda$)} 
In this section, we improve the state-of-the-art R-DQN method (Retrace($\lambda$)) with the proposed QM($\lambda$).  
We conduct such experiments on tasks from OpenAI Gym and Atari games, which is shown in Figure \ref{overall}. 
It is noticeable that the performance of DQN is reported as a baseline in Figure \ref{overall}. 

\textbf{Performance on OpenAI.}
We validate the effectiveness of QM($\lambda$) with two measurements on four representative OpenAI tasks, \emph{i.e.}, Cartpole-v1, Cartpole-v2, Copy and Pixelcopter. 
The performance comparison is shown on top figures in Figure \ref{overall}. 
As shown in Figure \ref{overall}, R-DQNs with our QM($\lambda$) achieve the highest scores among these compared algorithms on these OpenAI tasks. 
It is noticeable that R-DQNs with our QM($\lambda$) achieve higher scores than the state-of-the-art method. 

\textbf{Performance on Atari.} 
We validate the effectiveness of the proposed QM($\lambda$) with two measurements on four representative Atari games, \emph{i.e.}, berzerk, hero, qbert, and seaquest. 
The performance comparison is reported on the bottom figures in Figure \ref{overall}. 
From Figure \ref{overall}, we can observe that our QM($\lambda$) can help R-DQN achieve the highest score among these compared algorithms on these Atari games. 
It can be noted that higher scores are achieved by the proposed QM($\lambda$), compared to Retrace($\lambda$).

Such experimental results validate that the proposed QM($\lambda$) can help R-DQN outperform the state-of-the-art method.

\vspace{-8pt}
\section{Conclusion and Future Work}\label{sec:6}
In this paper, we propose an R-DQN framework. 
As compared to previous works, our R-DQN framework is able to combine general return-based algorithms with DQN.
Under the R-DQN framework, we propose a strategy to reasonably benefit from off-policy returns on near on- and off-policy cases. 
In order to qualitatively classify these two cases, we present two qualitative measurements and further give their bounds.  
The experimental results show that R-DQN algorithms in our R-DQN framework outperform the traditional DQN. 
The effectiveness of the proposed two measurements is validated by experiments. 
These two measurements also show their respective advantages on different kinds of tasks. 
It is indicated from the results that R-DQN with our QM($\lambda$) can outperform the state-of-the-art method. 

\textbf{Limitations and future works.} 
1) Note that the metric of $L_1$ we adopt for the formulations of measurements in Section \ref{section:iv-b} is quite simple, it could be challenging and interesting to use other metrics to measure the policy discrepancy, \emph{e.g.} Kullback-Leibler divergence. 
2) Despite our effects with reasonable definitions for near on- and off-policy cases in Section \ref{subsec:3.2}, such definitions are limited to one-step samples. More general definitions, which can take multi-step samples into account, are suggested as future work. 
3) Even though our derivation of measurement's bound in Section \ref{section:iv-b} is well-founded with a reasonable $\epsilon$ range assumption, however, it may not be applicable when there is an extreme demand for exploration due to the bounded $\epsilon$ range. Therefore, more general derivation processes are worthwhile in future work.


\ifCLASSOPTIONcaptionsoff
  \newpage
\fi

\vspace{-15pt}
\bibliographystyle{IEEEtran}
\bibliography{main}

\begin{thebibliography}{10}
\providecommand{\url}[1]{#1}
\csname url@samestyle\endcsname
\providecommand{\newblock}{\relax}
\providecommand{\bibinfo}[2]{#2}
\providecommand{\BIBentrySTDinterwordspacing}{\spaceskip=0pt\relax}
\providecommand{\BIBentryALTinterwordstretchfactor}{4}
\providecommand{\BIBentryALTinterwordspacing}{\spaceskip=\fontdimen2\font plus
\BIBentryALTinterwordstretchfactor\fontdimen3\font minus
  \fontdimen4\font\relax}
\providecommand{\BIBforeignlanguage}[2]{{%
\expandafter\ifx\csname l@#1\endcsname\relax
\typeout{** WARNING: IEEEtran.bst: No hyphenation pattern has been}%
\typeout{** loaded for the language `#1'. Using the pattern for}%
\typeout{** the default language instead.}%
\else
\language=\csname l@#1\endcsname
\fi
#2}}
\providecommand{\BIBdecl}{\relax}
\BIBdecl

\bibitem{sutton1998reinforcement}
R.~S. Sutton and A.~G. Barto, \emph{Reinforcement learning: An
  introduction}.\hskip 1em plus 0.5em minus 0.4em\relax MIT press, 1998.

\bibitem{SuttonMSM99}
R.~S. Sutton, D.~A. McAllester, S.~P. Singh, and Y.~Mansour, ``Policy gradient
  methods for reinforcement learning with function approximation,'' in
  \emph{NIPS}, 2000, pp. 1057--1063.

\bibitem{MnihKSRVBGRFOPB15}
V.~Mnih, K.~Kavukcuoglu, D.~Silver, A.~A. Rusu, J.~Veness, M.~G. Bellemare,
  A.~Graves, M.~A. Riedmiller, A.~Fidjeland, G.~Ostrovski, S.~Petersen,
  C.~Beattie, A.~Sadik, I.~Antonoglou, H.~King, D.~Kumaran, D.~Wierstra,
  S.~Legg, and D.~Hassabis, ``Human-level control through deep reinforcement
  learning,'' \emph{Nature}, vol. 518, no. 7540, pp. 529--533, 2015.

\bibitem{Lin92}
L.~J. Lin, ``Self-improving reactive agents based on reinforcement learning,
  planning and teaching,'' \emph{Machine Learning}, vol.~8, no. 3-4, pp.
  293--321, 1992.

\bibitem{MunosSHB16}
R.~Munos, T.~Stepleton, A.~Harutyunyan, and M.~G. Bellemare, ``Safe and
  efficient off-policy reinforcement learning,'' in \emph{NIPS}, 2016, pp.
  1054--1062.

\bibitem{WangBHMMKF16}
Z.~Wang, V.~Bapst, N.~Heess, V.~Mnih, R.~Munos, K.~Kavukcuoglu, and
  N.~de~Freitas, ``Sample efficient actor-critic with experience replay,'' in
  \emph{ICLR}, 2017.

\bibitem{gruslys2017reactor}
A.~Gruslys, W.~Dabney, M.~G. Azar, B.~Piot, M.~Bellemare, and R.~Munos, ``The
  reactor: A fast and sample-efficient actor-critic agent for reinforcement
  learning,'' in \emph{ICLR}, 2018.

\bibitem{Watkins1992}
C.~J. Watkins and P.~Dayan, ``Q-learning,'' \emph{Machine learning}, vol.~8,
  no. 3-4, pp. 279--292, 1992.

\bibitem{PengW96}
J.~Peng and R.~J. Williams, ``Incremental multi-step q-learning,''
  \emph{Machine Learning}, vol.~22, no. 1-3, pp. 283--290, 1996.

\bibitem{vanHasselt11}
H.~P. van Hasselt, ``Insights in reinforcement learning : formal analysis and
  empirical evaluation of temporal-difference learning algorithms,'' Ph.D.
  dissertation, Utrecht University, Netherlands, 2011.

\bibitem{HarutyunyanBSM16}
A.~Harutyunyan, M.~G. Bellemare, T.~Stepleton, and R.~Munos, ``Q({\(\lambda\)})
  with off-policy corrections,'' in \emph{ALT}, 2016, pp. 305--320.

\bibitem{BartoD93}
A.~G. Barto and M.~O. Duff, ``Monte carlo matrix inversion and reinforcement
  learning,'' in \emph{NIPS}, 1994, pp. 687--694.

\bibitem{GuLGTL16}
S.~Gu, T.~P. Lillicrap, Z.~Ghahramani, R.~E. Turner, and S.~Levine, ``Q-prop:
  Sample-efficient policy gradient with an off-policy critic,'' in \emph{ICLR},
  2017.

\bibitem{GuLTGSL17}
S.~Gu, T.~Lillicrap, R.~E. Turner, Z.~Ghahramani, B.~Sch{\"{o}}lkopf, and
  S.~Levine, ``Interpolated policy gradient: Merging on-policy and off-policy
  gradient estimation for deep reinforcement learning,'' in \emph{NIPS}, 2017,
  pp. 3846--3855.

\bibitem{PengW93}
J.~Peng and R.~J. Williams, ``Efficient learning and planning within the dyna
  framework,'' \emph{Adaptive Behavior}, vol.~1, no.~4, pp. 437--454, 1993.

\bibitem{PrecupSS00}
D.~Precup, R.~S. Sutton, and S.~P. Singh, ``Eligibility traces for off-policy
  policy evaluation,'' in \emph{ICML}, 2000, pp. 759--766.

\bibitem{SilverHMGSDSAPL16}
D.~Silver, A.~Huang, C.~J. Maddison, A.~Guez, L.~Sifre, G.~van~den Driessche,
  J.~Schrittwieser, I.~Antonoglou, V.~Panneershelvam, M.~Lanctot, S.~Dieleman,
  D.~Grewe, J.~Nham, N.~Kalchbrenner, I.~Sutskever, T.~P. Lillicrap, M.~Leach,
  K.~Kavukcuoglu, T.~Graepel, and D.~Hassabis, ``Mastering the game of go with
  deep neural networks and tree search,'' \emph{Nature}, vol. 529, no. 7587,
  pp. 484--489, 2016.

\bibitem{LillicrapHPHETS15}
T.~P. Lillicrap, J.~J. Hunt, A.~Pritzel, N.~Heess, T.~Erez, Y.~Tassa,
  D.~Silver, and D.~Wierstra, ``Continuous control with deep reinforcement
  learning,'' in \emph{ICLR}, 2016.

\bibitem{SchulmanLAJM15}
J.~Schulman, S.~Levine, P.~Abbeel, M.~I. Jordan, and P.~Moritz, ``Trust region
  policy optimization,'' in \emph{ICML}, 2015, pp. 1889--1897.

\bibitem{HasseltGS16}
H.~van Hasselt, A.~Guez, and D.~Silver, ``Deep reinforcement learning with
  double q-learning,'' in \emph{AAAI}, 2016, pp. 2094--2100.

\bibitem{WangSHHLF16}
Z.~Wang, T.~Schaul, M.~Hessel, H.~van Hasselt, M.~Lanctot, and N.~de~Freitas,
  ``Dueling network architectures for deep reinforcement learning,'' in
  \emph{ICML}, 2016, pp. 1995--2003.

\bibitem{SchaulQAS15}
T.~Schaul, J.~Quan, I.~Antonoglou, and D.~Silver, ``Prioritized experience
  replay,'' in \emph{ICLR}, 2016.

\bibitem{HesselMHSODHPAS18}
M.~Hessel, J.~Modayil, H.~van Hasselt, T.~Schaul, G.~Ostrovski, W.~Dabney,
  D.~Horgan, B.~Piot, M.~G. Azar, and D.~Silver, ``Rainbow: Combining
  improvements in deep reinforcement learning,'' in \emph{AAAI}, 2018, pp.
  3215--3222.

\bibitem{Lin93}
L.~J. Lin, ``Scaling up reinforcement learning for robot control,'' in
  \emph{ICML}, 1993, pp. 182--189.

\bibitem{Brockman}
G.~Brockman, V.~Cheung, L.~Pettersson, J.~Schneider, J.~Schulman, J.~Tang, and
  W.~Zaremba, ``Openai gym,'' \emph{arXiv preprint arXiv:1606.01540}, 2016.

\bibitem{AnschelBS17}
O.~Anschel, N.~Baram, and N.~Shimkin, ``Averaged-dqn: Variance reduction and
  stabilization for deep reinforcement learning,'' in \emph{ICML}, 2017, pp.
  176--185.

\end{thebibliography}

\end{document}